# ADAB: Arabic Dataset for Automated Politeness Benchmarking - A Large-Scale Resource for Computational Sociopragmatics

Hend Al-Khalifa[1], Nadia Ghezaiel[2], Maria Bounnit[3], Hend Hamed Alhazmi[4], Noof Abdullah Alfear[1], Reem Fahad Alqifari[1], Ameera Masoud Almasoud[1] and Sharefah Al-Ghamdi[1]

[1]College of Computer and Information Sciences, King Saud University, [2]College of Computer Science and Engineering, University of Hail, [3]Cadi Ayyad University, [4]Saudi Center of Philosophy and Ethics
[1]Riyadh, Saudi Arabia, [2]Hail, Saudi Arabia, [3]Marrakesh, Morocco, [4]Jeddah, Saudi Arabia
[1]{hendk|nalfear|ralgifary|ammalmasoud|sharefah}@ksu.edu.sa, {[2]ghezaielnadia.ing|[3]mariabounnit|
[4]hend.hamed.w}@gmail.com

**Abstract**

The growing importance of culturally-aware natural language processing systems has led to an increasing demand for resources that capture sociopragmatic phenomena across diverse languages. Nevertheless, Arabic-language resources for politeness detection remain severely under-explored, despite the rich and complex politeness expressions deeply embedded in Arabic communication. In this paper, a new annotated Arabic dataset, called ADAB/أدب (Arabic Politeness Dataset), was generated and carefully collected from four diverse online platforms including social media, e-commerce, and customer service domains, encompassing both Modern Standard Arabic (MSA) and multiple dialectal varieties (Gulf, Egyptian, Levantine, and Maghrebi). This dataset has undergone a thorough annotation process guided by Arabic linguistic traditions and contemporary pragmatic theory, resulting in three-way politeness classifications: polite, impolite, and neutral. The generated dataset contains 10,000 samples with detailed linguistic feature annotations across 16 politeness categories, achieving substantial inter-annotator agreement ($\kappa = 0.703$). A comprehensive benchmarking of this dataset was conducted utilizing 40 model configurations spanning traditional machine learning (12 models), transformer-based architecture (10 models), and large language models (18 configurations), thereby effectively demonstrating its practical utility and inherent challenges. This generated resource aims to bridge the gap in Arabic sociopragmatic NLP and encourage further research into politeness-aware applications for the Arabic language.

**Keywords:** Politeness Classification, Sociopragmatics, Language Resource, Arabic Natural Language Processing

## 1. Introduction

Politeness is a fundamental aspect of human communication that shapes social interactions across all languages and cultures. While the universality of politeness has been debated (Brown et al., 1987), its linguistic realization varies significantly across languages, reflecting cultural norms, social hierarchies, and pragmatic conventions. Arabic, spoken by over 400 million people across diverse geographical and cultural contexts (Darwish et al., 2021), exhibits particularly rich and complex politeness phenomena, including elaborate greeting formulas, honorifics, indirect speech acts, and religiously-informed expressions of courtesy.

Despite this richness, computational linguistic research on Arabic politeness is rare. Existing work on politeness classification and generation has focused almost exclusively on English with limited extensions to other languages (Priya et al., 2024). This gap is particularly problematic given the growing importance of Arabic in digital communication and the need for culturally-aware natural language processing systems that can navigate social norms appropriately.

Arabic presents unique challenges for politeness modeling. Its diglossia, the coexistence of Modern Standard Arabic (MSA) and numerous dialectal varieties, creates variation in politeness expression across registers. The language's morphological complexity, where politeness markers may be embedded in verb conjugations, pronoun forms, or lexical choices, requires sophisticated understanding beyond surface-level patterns. Additionally, Arabic's right-to-left script and distinct grammatical structures necessitate careful consideration in model development.

To address these challenges, we introduce ADAB/أدب, a dataset containing 10,000 Arabic texts annotated with three-way politeness labels: polite, impolite, and neutral. This dataset represents the first large-scale effort to computationally model Arabic politeness. Our research addresses the following questions:

- **RQ1**: How well can contemporary NLP models, ranging from traditional machine learning (ML) to large language models (LLMs), identify politeness levels in Arabic text?
- **RQ2**: What are the relative strengths of different modeling approaches (traditional ML, transformer-based, and LLMs) for capturing Arabic politeness phenomena?
- **RQ3**: What specific aspects of Arabic politeness pose the greatest challenges for automated classification?

Thus, our contributions are as follows:

1. **ADAB Dataset**: The first large-scale Arabic politeness dataset with 10,000 annotated examples across three politeness categories, providing a critical resource for Arabic NLP research.

2. **Comprehensive Evaluation**: Systematic assessment of 40 model configurations spanning three paradigms (12 traditional ML models, 10 transformer-based models, and 18 LLM configurations), establishing baseline performances for Arabic politeness classification.
3. **Linguistic Analysis**: Detailed error analysis revealing key challenges including neutral classification bias, implicit marker detection difficulties, context-dependent interpretation of religious expressions, and dialectal variation handling.

The critical importance of ADAB extends beyond academic interest. For the hundreds of millions of Arabic speakers worldwide, having NLP systems that understand and respect politeness norms is essential for effective human-computer interaction. Customer service chatbots, educational applications, social media moderation tools, and translation systems all benefit from politeness awareness. Furthermore, as Arabic digital content continues to grow exponentially, tools for analyzing social dynamics, detecting harassment, and facilitating respectful communication become increasingly vital.

The remainder of this paper is organized as follows: Section 2 describes our dataset construction methodology. Section 3 presents our experimental setup and model evaluation. Section 4 analyzes results. Section 5 discusses error analysis and challenges. Section 6 reviews related work, and Sections 7, 8 and 9 conclude with future directions, limitations and ethical considerations.

## 2. ADAB Dataset Construction

### 2.1 Motivation

Our goal is to construct a large-scale, high-quality Arabic politeness dataset that captures the linguistic and cultural richness of Arabic politeness phenomena. Arabic politeness is deeply intertwined with Islamic values, social hierarchies, and regional variations, making it a complex sociopragmatic phenomenon. Following the framework established for English (Danescu-Niculescu-Mizil et al., 2013), we focus on naturally occurring texts where speakers navigate social interactions, employing various politeness strategies or, conversely, exhibiting impoliteness.

Unlike previous multilingual politeness work that relied on translation (which can lose cultural nuances), we collect native Arabic content to preserve authentic linguistic markers. We adopt a three-way classification scheme: polite, impolite, and neutral, recognizing that many utterances fall into an intermediate category where explicit politeness strategies are absent, but the text is not overtly impolite.

### 2.2 Data Sources

We collected 10,000 Arabic texts from four publicly available datasets, ensuring diversity in domain, register, and communicative purpose:

- **YouTube Comments (~70K)**: Comments on Arabic content creators' videos[1], representing informal, spontaneous reactions ranging from appreciative responses to critical feedback. This platform captures colloquial Arabic across various dialects.
- **Product Reviews (~6K)**: Customer reviews on the Shein e-commerce platform[2], containing evaluations of products and services. These texts often exhibit politeness strategies when praising products or diplomatic criticism when expressing dissatisfaction.
- **Twitter/X Posts (~20K)**: The ArSAS[3] dataset a public tweet in Arabic covering social, cultural, and political discourse. Twitter's character limit encourages concise expression where politeness markers may be amplified or omitted.
- **Application Reviews (~67K)**: User feedback on a mobile banking application[4], representing customer-to-business communication where politeness norms may differ from peer-to-peer interaction.

To ensure annotation quality and focus on genuine politeness phenomena, data cleaning removed items containing critical political content or direct references to public figures, profanity or highly explicit insults, unacceptable criticism of religious beliefs, or single-word/semantically incoherent utterances. This multi-domain approach captures varied politeness phenomena from direct peer communication to customer feedback contexts where power dynamics influence linguistic choices.

### 2.3 Annotation Framework

#### 2.3.1 Theoretical Framework

Our annotation framework synthesizes Arabic linguistic traditions with contemporary pragmatic theory. Classical Arabic rhetoric emphasized gentle speech (القول اللين) and the good word (الكلمة الطيبة), concepts rooted in Quranic verses and Prophetic traditions. We integrate these culturally-grounded notions with (Lakoff, 1973) politeness maxims and (Leech, 2016) politeness principles,

---
[1] https://www.kaggle.com/datasets/farisalahmdi/arabic-youtube-comments-by-khalaya
[2] https://huggingface.co/datasets/Ruqiya/Arabic_Reviews_of_SHEIN + https://www.kaggle.com/datasets/omaima52/arabic-shein-reviews-emotion-detection
[3] https://huggingface.co/datasets/arbml/ArSAS
[4] https://www.kaggle.com/datasets/mohamedalisalama/arabic-companies-reviews-for-sentiment-analysis

adapted for Arabic's morphosyntactic and cultural specificity.

### 2.3.2 Politeness and Impoliteness Categories

Rather than relying solely on surface-level linguistic features, we ground our annotation in broader communicative categories that reflect Arabic cultural norms:

*Politeness Categories:*

1. Appreciation, Admiration, and Love (Using language that inherently expresses deep respect.)
2. Asking for permission and courtliness
3. Congratulations
4. Greetings
5. Hospitality and Generosity
6. Gratitude and Thanks
7. Respect

*Impoliteness Categories:*

8. Accusation
9. Criticism (harsh or unnecessarily aggressive)
10. Discrimination, racism, and sectarian attacks
11. Disparagement
12. Insult
13. Sarcasm (when used to demean)
14. Threat
15. Verbal violence

*Both Categories:*

16. Prayers: In Arabic, there is a fundamental distinction between *Du'ā' Lak* (a prayer for someone, invoking good) and *Du'ā' 'Alayk* (a prayer against someone, invoking harm). The same linguistic form is used for blessings and curses.

These categories provide the conceptual foundation for our three-way classification and help annotators identify the communicative intent behind linguistic choices. Importantly, the 16 categories serve as an annotation guide rather than a 16-way classification task; the primary classification remains three-way (polite, impolite, and neutral), while the fine-grained categories offer additional sociopragmatic richness that can support future research on specific politeness strategies.

### 2.4 Annotation Process

#### 2.4.1 Annotator Selection and Training

Two annotators with Ph.D. degrees in Arabic linguistics were recruited to ensure deep understanding of Arabic morphosyntax, pragmatics, and dialectal variation. Both annotators are native Arabic speakers familiar with multiple Arabic dialects represented in the dataset (Gulf, Egyptian, Levantine, and Maghrebi varieties). They are also part of the project team.

Prior to annotation, annotators underwent extensive training:

1. Study of classical and modern Arabic theories of politeness.
2. Review of the 16 politeness and impoliteness categories with examples.
3. Practice annotation on 100 pilot examples with discussion.
4. Iterative refinement of annotation guidelines based on ambiguous cases.

#### 2.4.2 Annotation Workflow

The annotation proceeded in systematic stages:

*Stage 1: Data Preprocessing*

- Identification and removal of duplicate texts.
- Removal of emojis.
- Exclusion of overtly offensive content violating community standards.

*Stage 2: Politeness Classification*

Each text was assigned one of three labels based on both the categories and observable linguistic features:

- **Polite:** Text where the author uses words and makes an evident effort to demonstrate kindness toward the interlocutor or public. This includes expressions of appreciation, gratitude, prayers, respect, courtliness, or hospitality. The text employs one or more politeness strategies, showing consideration for the addressee.

- **Neutral:** Text where the writer adheres to fundamental norms of expression, conveying their perspective in a generally acceptable manner. It may employ evaluative words like "good" or "bad" at a standard level of expression. The text lacks explicit politeness markers but does not include impoliteness.

- **Impolite:** Text employing inappropriate language, directly attacking the audience, offering criticism in a needlessly harsh manner, or exhibiting verbal violence, discrimination, disparagement, accusations, threats, or demeaning sarcasm. The text contains rudeness, aggression, or other forms of disrespect.

*Stage 3: Keyword Extraction*

Identification and extraction of politeness-related keywords and phrases.

This process created a rich feature matrix documenting not only the overall politeness classification but also the specific communicative strategies and linguistic realizations present in each text.

#### 2.4.3 Collaborative Annotation Protocol

To ensure consistency, annotators worked in synchronized daily batches:

1. Each annotator independently labeled approximately 100 texts per day.

2. Annotators met via video call to review the day's annotations, discussing disagreements and edge cases.
3. Process repeated daily until 10,000 texts completed.
4. Complete revision of all annotated texts by both annotators for final consistency check.

This iterative approach took **9 months**, balancing annotation quality with efficiency.

### 2.5 Annotation Challenges and Solutions

Arabic politeness annotation presented unique challenges:

1. **Ideological Content**: Texts with strong religious sectarian, political partisan content was excluded to avoid conflating ideology with politeness.

2. **Context-Dependent Polysemy**: Words like سلامات ("well wishes" or sarcastic "good riddance") required careful contextual interpretation.

3. **Product Praise**: Generic positive adjectives in product reviews were distinguished from interpersonal politeness.

4. **Implicit Expressions**: Some impolite texts lacked explicit offensive keywords, requiring pragmatic inference.

5. **Intensifier Ambiguity**: Amplifiers could enhance both positive and negative meanings, requiring holistic text interpretation.

6. **Morphological Complexity**: Arabic's rich derivational morphology (منك "from you-sg." vs. منكم "from you-pl.honorific") required attention to subtle distinctions.

7. **Dialectal Diversity**: Texts included Gulf, Egyptian, Levantine, and Maghrebi dialects, each with distinct politeness norms. Annotators documented dialectal features and interpreted them within their respective conventions.

8. **Political and Religious Content:** Texts with strong political or sectarian content were excluded during data cleaning to avoid conflating ideology with politeness expression.

### 2.6 Inter-Annotator Agreement

To assess annotation quality, we computed Cohen's Kappa coefficient on a stratified sample of 2000 texts. Our annotators achieved κ = 0.703, indicating substantial agreement (Landis & Koch, 1977). This score demonstrates that despite the inherent subjectivity of politeness judgments and the complexity of Arabic's linguistic features, annotators maintained consistent interpretations.

This agreement level is comparable to other subjective NLP tasks and slightly higher than the average reported in multilingual politeness studies (Srinivasan & Choi, 2022): average pairwise Spearman ρ = 0.46 for native annotations). The higher agreement may reflect our annotators' linguistic expertise and the extensive training process.

### 2.7 Quality Control Measures

Beyond inter-annotator agreement, we implemented several quality assurance steps:

- **Consistency Checks**: A complete revision for all annotated texts by both annotators, confirming agreement rates remained stable throughout the project.
- **Edge Case Review**: Texts with disagreement were flagged for joint discussion and consensus labeling.
- **Feature Validation**: The politeness and impoliteness categories were reviewed by three additional Arabic linguists, independent of the annotation process, who confirmed their validity as reliable markers.

### 2.8 Final Dataset Statistics

The final ADAB dataset contains:

- 10,000 Arabic texts with politeness labels.
- 4 source domains (2,500 texts each).
- 3 politeness classes: Polite, Neutral, Impolite.
- 16 linguistic politeness and impoliteness categories annotated per text.
- Average text length (words): 13.90 Words.
- Class distribution: 19.12% polite, 70.72% neutral, 10.16% impolite.

Figure 1 shows the distribution of politeness labels across the four domains, revealing domain-specific patterns in politeness expression.

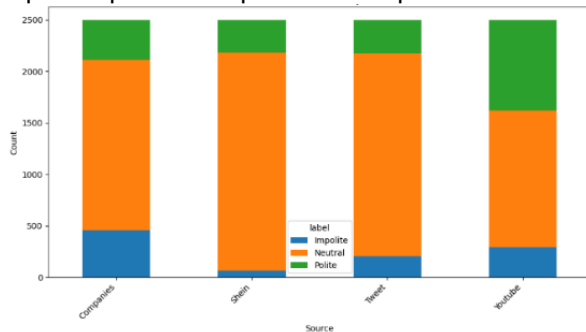

Figure 1: Distribution of politeness labels across the four domains

## 3. Evaluation Methodology

This comprehensive evaluation benchmark investigates the effectiveness of diverse classification models for politeness detection. It encompasses three main groups of approaches: (1) traditional machine learning (ML) models, (2) transformer-based architectures, and (3) large language models (LLMs).

The benchmark begins with three traditional classifiers (Logistic Regression, SVM, and XGBoost) trained using both sparse lexical features (TF-IDF) and dense word embeddings (FastText, GloVe, and Word2Vec) to establish baseline performance. It then extends to fine-tuned transformer-based models, divided into Arabic-specific variants (AraBERTv02, MARBERT, and CAMeLBERT-mix) and multilingual models (XLM-RoBERTa, mDeBERTa, RemBERT, InfoXLM, and mBERT), allowing a detailed comparison of language-specific versus cross-lingual pre-training effects. Finally, the benchmark evaluates LLMs under zero-shot or/and few-shot settings, including proprietary systems (GPT-4 mini, and Claude -sonnet-4.5), Arabic open-weights models (Fanar-1-9B, and ALLaM-7B-Instruct-preview), and multilingual LLMs (DeepSeek V3, and GPT OSS 20B) (all models are from HuggingFace[5]). Together, these models form a robust benchmark for analyzing how modeling strategies, pre-training paradigms, and representational capacities influence Arabic politeness classification.

| Traditional ML Models | Accuracy | Macro-F1 | Transformer-Based Models | Accuracy | Macro-F1 | LLMs | Accuracy | Macro-F1 |
|---|---|---|---|---|---|---|---|---|
| LR (TF-IDF) | 0.79 | 0.67 | AraBERTv02 | 0.8984 | 0.8296 | Fanar-1-9B (3-shot) | 0.7485 | 0.2854 |
| LR (Word2Vec) | 0.71 | 0.62 | MARBERT | **0.9119** | **0.8582** | ALLaM-7B-Instruct-preview (3-shot) | 0.3000 | 0.1667 |
| LR (GloVe) | 0.67 | 0.57 | CAMeL BERT-mix | 0.8929 | 0.8193 | Deepseek V3 (0-shot) | 0.6534 | 0.402 |
| LR (FastText) | 0.71 | 0.63 | mBERT-cased | 0.8596 | 0.7309 | Deepseek V3 (1-shot) | 0.5687 | 0.368 |
| SVM(TF-IDF) | **0.84** | **0.71** | mBERT-uncased | 0.8625 | 0.7314 | Deepseek V3 (3-shot) | 0.5747 | 0.372 |
| SVM (Word2Vec) | 0.78 | 0.66 | XLM-RoBERTa base | 0.8720 | 0.7365 | GPT OSS 20b (0-shot) | 0.818 | 0.661 |
| SVM (GloVe) | 0.77 | 0.61 | XLM-RoBERTa large | 0.8889 | 0.8150 | GPT OSS 20b (1-shot) | 0.8217 | 0.667 |
| SVM (FastText) | 0.81 | 0.68 | mDeBERTa v3 | 0.8959 | 0.8252 | GPT OSS 20b (3-shot) | 0.8344 | 0.678 |
| XGBoost (TF-IDF) | 0.81 | 0.67 | InfoXLM | 0.8616 | 0.7521 | GPT-4 mini (0-shot) | 0.6150 | 0.5746 |
| XGBoost (Word2Vec) | 0.84 | 0.68 | RemBERT | 0.8830 | 0.7891 | GPT-4 mini (1-shot) | 0.6579 | 0.6041 |
|  |  |  | **LLMs** | **Accuracy** | **Macro-F1** |  |  |  |
| XGBoost (GloVe) | 0.82 | 0.63 | Fanar-1-9B (0-shot) | 0.3000 | 0.1667 | GPT-4 mini (3-shot) | 0.8167 | 0.6527 |
|  |  |  | ALLaM-7B-Instruct-preview (0-shot) | 0.5896 | 0.5228 | Claude -sonnet-4.5 (0-shot) | 0.7485 | 0.2854 |
|  |  |  | Fanar-1-9B (1-shot) | 0.8000 | 0.3137 | Claude -sonnet-4.5 (1-shot) | 0.7485 | 0.2854 |
| XGBoost (FastText) | 0.84 | 0.69 | ALLaM-7B-Instruct-preview (1-shot) | 0.5936 | 0.5467 | Claude -sonnet-4.5 (3-shot) | **0.8426** | **0.7131** |

Table 1: Models Performance Summary

To ensure fair and comprehensive evaluation, the benchmark employs multiple metrics, Accuracy, micro-F1, and macro-F1, accounting for the dataset's class imbalance, where Neutral examples dominate. While Accuracy captures overall correctness, it may overstate performance in imbalanced scenarios. Thus, F1-based metrics were used to provide a more balanced assessment: micro-F1 reflects the global effectiveness across all classes, whereas macro-F1 gives equal importance to each class, emphasizing performance on minority categories such as Polite and Impolite.

## 4. Results

The benchmark results in Table 1 demonstrate that transformer-based models outperform traditional and LLMs in Arabic politeness detection. Among traditional methods, SVM (TF-IDF) achieved the best performance (Accuracy = 0.84, Macro-F1 = 0.71). MARBERT, ranked highest overall (Accuracy = 0.9119, Macro-F1 = 0.8582), confirming the advantage of language-specific fine-tuning. In contrast, LLMs showed mixed results, few-shot settings sometimes improved performance, with Claude -sonnet-4.5 (1-shot) and GPT OSS 20b (3-shot) performing best, while Arabic open-source models trailed behind. Overall, fine-tuned transformers demonstrated the most reliable and robust performance across metrics.

---
[5] https://huggingface.co/

Table 2 presents the top five performing models in the politeness classification task. MARBERT also maintained strong performance across all three classes: impolite, neutral, and polite, showing its ability to capture subtle differences in tone. Although AraBERTv02 and mDeBERTa v3 achieved comparable overall scores, their per-class F1 values reveal different behavior:

AraBERTv02 performed slightly better in recognizing impolite instances. All models performed consistently well in detecting neutral and polite expressions, with only minor variations among them. However, MARBERT showed a clear advantage in the impolite class, where other models tended to struggle. This indicates that MARBERT's pretraining on large-scale Arabic social media data, which includes dialectal and informal expressions, helps it better recognize impolite or emotionally charged language. Overall, these results emphasize the strength of Arabic-specific BERT models, particularly MARBERT, in handling nuanced Arabic language use across different politeness levels.

| Model | Accuracy | Macro-F1 | F1 Impolite/ Neutral/ Polite |
|---|---|---|---|
| **MARBERT** | **0.9119** | **0.8582** | **0.81/ 0.94/ 0.82** |
| AraBERTv02 | 0.8984 | 0.8296 | 0.74/ 0.93/ 0.82 |
| mDeBERTa v3 | 0.8959 | 0.8252 | 0.72/ 0.93/ 0.82 |
| CAMeL BERT-mix | 0.8929 | 0.8193 | 0.73/ 0.93/ 0.80 |
| XLM-RoBERTa large | 0.8889 | 0.8150 | 0.71/ 0.93/ 0.81 |

Table 2: Top-5 Models Detailed Metrics

To further analyze the behavior of the different modeling paradigms, a statistical comparison was conducted at two levels: (i) across the three main model families: traditional ML models, transformer-based models, and LLMs; and (ii) among the five top-performing individual models. Mean Accuracy and Macro-F1 scores were computed among the three models' families from the results reported in Table 1, revealing a clear performance hierarchy in which transformer-based models consistently outperform both traditional classifiers and LLMs. This superiority highlights the effectiveness of contextualized Arabic embeddings and domain-specific pretraining in capturing subtle pragmatic cues. The average Macro-F1 scores for traditional machine learning models, transformer-based models, and LLMs were 0.65, 0.79, and 0.56, respectively, which proves that transformer-based models lead by a substantial margin. This superiority can be attributed to the distinctive strengths of each family. Traditional ML models, although competitive when paired with strong lexical features like TF-IDF or FastText, rely on fixed representations that lack contextual sensitivity. As a result, they struggle to distinguish subtle pragmatic shifts, such as polite mitigation versus passive aggression, that depend on surrounding discourse. LLMs, despite their vast pretraining, underperform in this task because they were not fine-tuned on politeness-specific or Arabic-focused data. Their zero-shot and few-shot outputs reveal a tendency toward generic or majority-class predictions, indicating insufficient adaptation to sociolinguistic nuance. In contrast, transformer-based models fine-tuned on the task exhibit dedicated alignment with Arabic politeness cues. Their contextualized embeddings allow them to interpret meaning relative to surrounding words, rather than in isolation. Moreover, Arabic-specific transformers such as MARBERT and AraBERTv02 benefit from pretraining on dialectal, informal, and socially expressive Arabic, which equips them to recognize pragmatic markers like indirectness, sarcasm, or softened criticism. This combination of contextual encoding and linguistically-relevant pretraining explains why transformer-based models form the top-performing family in this benchmark.

A closer comparison of the five best-performing models shows that MARBERT achieved the highest Macro-F1 score (0.8582). Although the performance differences among the top models appear relatively small (approximately two to four percentage points), they represent a consistent and practically meaningful advantage for MARBERT. Class-wise F1 comparisons across Impolite, Neutral, and Polite categories (Table 2), indicate that all models perform strongly on Neutral (≈ 0.93) and Polite (≈ 0.81–0.82) classes, whereas Impolite instances remain more challenging. Scores for this class range from 0.71 (XLM-RoBERTa Large) to 0.81 (MARBERT), reflecting the inherent difficulty of detecting sarcasm and hostile tone in Arabic. MARBERT's superior performance in this category further underscores its robustness in modeling emotionally charged and dialectal expressions. Overall, the observed performance gaps (Δ Macro-F1), as shown in Table 3, provide consistent evidence that Arabic-specific pretraining yields practically meaningful improvements.

| Comparison | ΔMacro-F1 | Interpretation |
|---|---|---|
| MARBERT – AraBERTv02 | +0.0286 | Practically significant |
| MARBERT – mDeBERTa v3 | +0.0330 | Practically significant |
| MARBERT – CAMeL BERT-mix | +0.0389 | Moderate difference |
| MARBERT – XLM-RoBERTa large | +0.0432 | Clear improvement |

Table 3: Practical Difference in Macro-F1 Between Top Five Models

## 5. Error Analysis

To better understand the types of misclassifications produced by different model families, we conducted an error pattern analysis across polarity transitions. As shown in Table 4, the most dominant error across all systems is a neutral bias, wherein both polite and impolite utterances are frequently misclassified as neutral. This tendency is particularly pronounced in LLMs (up to 48% of errors) and traditional ML models (33–43%), while transformer-based models reduce this phenomenon to approximately 24%. Distinct secondary biases also emerge across model types: LLMs show an over-politeness tendency (18% Neutral→Polite), whereas ML models are more prone to over-assigning impoliteness (7.5% Neutral→Impolite). The most critical error, cross-confusion between Polite and Impolite classes, is minimal in transformer-based models (0.6%), moderate in ML (3.6%), and disproportionately high in LLMs (11.3%). Overall, transformer-based models achieve the lowest average error rate (10.9%), confirming their robustness, while ML and especially LLMs underperform without task-specific fine-tuning.

| Error Type | Traditional ML Models | Transformer-Based Models | LLMs | Average |
|---|---|---|---|---|
| Polite → Neutral | 33.38% | 23.95% | 48.23% | 35.19% |
| Impolite → Neutral | 42.74% | 23.98% | 38.82% | 35.18% |
| Neutral → Impolite | 7.50% | 3.43% | 6.02% | 5.65% |
| Neutral → Polite | 8.03% | 2.59% | 18.37% | 9.66% |
| Impolite ↔ Polite | 3.56% | 0.59% | 11.32% | 5.16% |
| **Average** | 19.04% | 10.91% | 24.55% | 18.17% |

Table 4: Error Distribution Across Model Types

Also, a qualitative error analysis of MARBERT's misclassifications reveals several recurring sources of difficulty. The most frequent issue involved **Lexical Markers**, where the model failed to detect explicit cues of politeness or impoliteness, resulting in Neutral predictions. **Ambiguous cases**, particularly those containing religious or conventionally polite formulas with context-dependent meanings, further contributed to confusion. **Contextual cases**, such as exaggerated refusals or implied criticism, were often interpreted literally rather than pragmatically. Finally, **cultural and dialectal expressions**, especially idioms from regional varieties like Egyptian Arabic, were frequently misinterpreted due to limited dialectal coverage in pretraining. These patterns suggest that improving sensitivity to pragmatic markers, contextual cues, and dialectal variation is key to enhancing model robustness.

**Error Category 1: Lexical Markers** (32.2% of Polite and Impolite errors)

- Example 1: " فضيلة الشيخ صلاح بن محمد البدير إمام وخطيب بالمسجد النبوي الشريف يؤم المصلين غدا لصلاة الاستسقاء بالمسجد النبوي الشريف" - **"fadilat al-shaykh salah bin muhammad al-budeir, imam wa-khatib al-masjid al-nabawi al-sharif, yu'umm al-musallin ghadan li-salat al-istisqa bil-masjid al-nabawi al-sharif."** In English "His Eminence Sheikh Salah bin Muhammad Al-Badr, Imam and Preacher of the Prophet's Mosque, will lead the worshippers tomorrow in the Istisqa' prayer at the Prophet's Mosque."
- True: Polite | Predicted: Neutral
- Analysis: The model was unable to detect the lexical markers to label the segment as polite.
- Example 2: " اجتماعات القبائل. والأغاني. والحماقات الأخرى-. جميعها حطب حصار قطر" **"ijtimaat al-qaba'il, wal-aghani, wal-hamaqaat al-ukhra, jami'uha hatab hisar qatar."** In English ""Tribal meetings, songs, and other foolish acts, all of them are fuel for the siege of Qatar."
- True: Impolite. | Predicted: Neutral
- Analysis: The word *"wal-hamaqaat"* is intended to refer disapprovingly in the sentence, which the model seems not to detect.

**Error Category 2: Ambiguous Cases** (12.88% of mixed errors)

- *Example:*
Text:" سبحان الله عجزت اتقبل الإنسان" **Subhan'Allah ajazt ataqabbal al insan"** - In English "Glory be to God, I am unable to accept the human."
- True: Impolite | Predicted: Neutral
- Analysis: Direct criticism misclassified due to the use of the religious expression *"Subhan'Allah"* which is intended to express exclamation depending on the context (it can be polite or impolite).

**Error Category 3: Contextual cases** (7.36% of mixed errors)

- Example: "أنا لن أشاهد الحلقة حتى لو أعطيتني مليون دولار" "**ana lan ushahid al halqa hatta law a'taytani million dollar**". In English "I wouldn't watch the episode even if you gave me a million dollars."
- True: Impolite | Predicted: Neutral
- Analysis: This is a culturally contextual expression that should be classified as impolite because of its construction, leading it to be full of bragging.

**Error Category 4: Cultural/Dialectal Markers** (11.04% of mixed errors)

- Example: "الواحد اذا تفلسف يجيب العيد"-"**al wahid iza tafalsaf yjeeb al eid**". In English "When someone tries to act too smart, they mess things up."
- True: Impolite | Predicted: Neutral
- Analysis: An idiomatic expression misclassified due to the incomprehension of the whole expression and a literal meaning.

In summary, transformer-based models, especially Arabic-specific ones, were most effective for Arabic politeness detection. MARBERT performs best overall, highlighting the value of linguistically grounded pretraining and the need for broader dialectal and pragmatic modeling.

## 6. Related Work

Computational politeness research has advanced notably in the past decade but remains largely focused on English. Our discussion centers on three key areas: datasets and annotation frameworks, modeling approaches, and pragmatic-cultural factors.

**Datasets and Annotation Frameworks**: Early work by (Danescu-Niculescu-Mizil et al., 2013) pioneered computational politeness using Wikipedia and Stack Exchange data, establishing that politeness could be modeled through observable linguistic strategies. Recent efforts have expanded both scope and sophistication. TYDIP (Srinivasan & Choi, 2022) frames politeness as scalar rather than categorical across nine languages, enabling nuanced modeling of intensity and pragmatic signals, though Arabic was notably absent. Domain-specific datasets like PolitePEER (Bharti et al., 2024) separate civility from surface politeness markers in peer-review discourse, highlighting how norms vary across communicative contexts. For Arabic specifically, resources remain severely limited. (Al-Khalifa et al., 2024) contributed 500 labeled MSA tweets, but this small-scale effort underscores the absence of comprehensive resources. ADAB addresses this gap with 10,000 texts spanning multiple domains and dialectal varieties.

**Modeling Approaches**: The landscape has transformed from traditional machine learning through transformer-based fine-tuning to large language models. Arabic-specific PLMs like MARBERT and CAMeLBERT-mix capture dialectal and informal patterns more effectively than multilingual alternatives, developing sensitivity to pragmatic cues embedded in Arabic morphology and indirect speech acts (Al-Khalifa et al., 2024). Contemporary LLMs show promise, PolitePEER found instruction-tuned models (Mistral-7B, LLaMA-13B) can outperform task-specific baselines. However, evaluations reveal mixed results for Arabic, with considerable sensitivity to genre, dialect, and prompt design (Al-Khalifa et al., 2024). (Aljanaideh, 2025) found weak out-of-domain generalization in politeness tasks, indicating that capacity alone does not ensure robust cross-lingual performance. These findings motivate our systematic benchmark across traditional ML, fine-tuned transformers, and LLMs.

**Pragmatic and Cultural Considerations**: Politeness is deeply embedded in cultural norms and social hierarchies. (Aljanaideh, 2025) demonstrated that incorporating speech-act patterns improves generalization by addressing underlying interactional functions rather than surface features. (Chen et al., 2025) showed that seemingly similar politeness concepts across languages index fundamentally different social meanings, demonstrating that direct category transfer risks cultural misrepresentation. Even prompt politeness affects LLM performance differently across languages (Yin et al., 2024). For Arabic, with its Islamic values, regional variation, and diglossia, cultural grounding is essential.

ADAB directly addresses identified gaps by providing the first large-scale, multi-domain Arabic politeness dataset grounded in classical Arabic linguistic traditions and contemporary pragmatic theory, with comprehensive evaluation establishing baselines and revealing challenges specific to Arabic's morphological complexity and cultural expressions.

## 7. Conclusion

We presented ADAB, the first large-scale Arabic politeness dataset comprising 10,000 texts annotated with three-way politeness classifications across four diverse domains. Through comprehensive evaluation of 40 model configurations, we addressed three fundamental research questions about Arabic politeness detection.

Regarding RQ1, our findings demonstrate that contemporary NLP models achieve moderate to strong performance, with the best transformer-based model (MARBERT) reaching 91.19% accuracy and 85.82% Macro-F1. However, substantial performance variation exists, ranging from 59.36% to 91.19% accuracy across all models, revealing significant room for improvement. For RQ2, transformer-based models (average Macro-F1: 79%) substantially outperformed both traditional ML methods (65%) and LLMs (56%), with Arabic-specific pre-training providing clear advantages over multilingual or zero-shot approaches. Addressing RQ3, error analysis identified four critical challenges: neutral bias across all model types, difficulty detecting implicit politeness markers (32% of errors), limited handling of dialectal variation, particularly Egyptian Arabic (11% of errors), and context-dependent ambiguity in religious expressions and sarcasm (13% of errors).

The ADAB dataset will support applications in politeness classification, dialogue systems, sentiment analysis, and social media moderation. Future work will focus on expanding dialectal coverage with more balanced representation of Egyptian and Maghrebi varieties, incorporating conversational context and interlocutor metadata, and extending annotation to formal registers. We will release the dataset, guidelines, and benchmarks to advance Arabic sociopragmatics and cross-lingual politeness research.

## 8. Limitations

The ADAB dataset has several limitations that we acknowledge and discuss here to guide future work. First, significant class imbalance (70.72% neutral, 19.12% polite, 10.16% impolite) creates modeling challenges, as evidenced by consistent neutral bias across all systems. While this distribution reflects the natural prevalence of neutral language in online communication, it limits the stability of fine-grained politeness and impoliteness modeling. Preliminary experiments with class weighting and oversampling narrowed the performance gap but did not eliminate the neutral bias, suggesting that this tendency is partly an inherent property of the pragmatic task itself. Future work could explore stratified or balanced subsets for targeted analysis of minority classes.

Second, the dataset primarily represents informal user-generated content from social media, e-commerce, and customer service domains, limiting its generalizability to formal registers such as diplomatic communication, academic discourse, or formal correspondence, where politeness strategies may differ substantially.

Third, despite substantial inter-annotator agreement (κ = 0.703), politeness judgments remain inherently subjective and context-dependent. While our two expert annotators bring deep expertise in Arabic pragmatics and dialectal variation, expanding the annotator pool in future iterations would provide additional validation and enable more robust disagreement analysis across annotators with different dialectal backgrounds. We note that three additional Arabic linguists independently validated the 16 politeness and impoliteness categories (see Section 2.6), providing a form of external validation of the annotation framework.

Fourth, many texts lack conversational context and speaker relationship information, which are important for politeness interpretation. The current dataset operates at the utterance level, consistent with foundational work such as Danescu-Niculescu-Mizil et al. (2013) and TYDIP (Srinivasan & Choi, 2022). However, we recognize that politeness is fundamentally context-dependent and relational. A planned future version (ADAB v2) will incorporate conversational threads, reply chains, and speaker metadata where available from the source platforms.

Fifth, dialectal coverage is uneven, with Gulf Arabic and MSA predominating due to the geographic origin of several source platforms, while Egyptian and Maghrebi varieties are underrepresented. This partly explains model difficulties with these varieties, as discussed in the error analysis. Future iterations would benefit from targeted collection of underrepresented dialects.

Finally, our LLM evaluation used zero-shot and few-shot settings without extensive prompt engineering or fine-tuning, which may not represent optimal performance. Despite these limitations, ADAB provides a critical foundation for Arabic politeness research and culturally-aware NLP application.

## 9. Ethics Statement

This research complies with the LREC 2026 Ethical Guidelines for responsible language resource development. All data were collected from publicly available sources in accordance with platform terms of service. Profanity, hate speech, and politically sensitive content were removed during preprocessing, and all texts were anonymized to protect user privacy. Annotation was conducted by two Ph.D.-level members of the research team; no external annotators or compensation were involved. The dataset is released solely for research and educational purposes, with care taken to minimize potential bias or harm.

## 10. Language Resource References

- **ADAB (Arabic Politeness Dataset).** Developed in this study; 10,000 annotated Arabic texts across four domains. The dataset will be publicly released after paper acceptance for research use.
- **ArSAS Dataset**. Arabic tweets annotated for sentiment and social attitudes. https://huggingface.co/datasets/arbml/ArSAS
- **Arabic YouTube Comments by Khalaya.** User comments from Arabic YouTube channels.

- https://www.kaggle.com/datasets/farisalahmdi/arabic-youtube-comments-by-khalaya
- **Arabic Reviews of SHEIN.** E-commerce product reviews in Arabic. https://huggingface.co/datasets/Ruqiya/Arabic_Reviews_of_SHEIN
- **Arabic Companies Reviews.** User reviews of Arabic mobile banking and service apps. https://www.kaggle.com/datasets/mohamedalisalama/arabic-companies-reviews-for-sentiment-analysis

## 11. References

Aljanaideh, A. (2025). Speech Act Patterns for Improving Generalizability of Explainable Politeness Detection Models. In W. Che, J. Nabende, E. Shutova, & M. T. Pilehvar (Eds.), *Findings of the Association for Computational Linguistics: ACL 2025* (pp. 18945–18954). Association for Computational Linguistics. https://doi.org/10.18653/v1/2025.findings-acl.970

Al-Khalifa, H., Ghezaiel, N., & Bounnit, M. (2024). Analyzing Politeness in Arabic Tweets: A Preliminary Study. In M. Abbas & A. A. Freihat (Eds.), *Proceedings of the 7th International Conference on Natural Language and Speech Processing (ICNLSP 2024)* (pp. 352–359). Association for Computational Linguistics. https://aclanthology.org/2024.icnlsp-1.36/

Bharti, P. K., Navlakha, M., Agarwal, M., & Ekbal, A. (2024). PolitePEER: Does peer review hurt? A dataset to gauge politeness intensity in the peer reviews. *Language Resources and Evaluation*, *58*(4), 1291–1313. https://doi.org/10.1007/s10579-023-09662-3

Brown, P., Levinson, S. C., & Gumperz, J. J. (1987, February 27). *Politeness: Some Universals in Language Usage*. Cambridge Aspire Website; Cambridge University Press. https://doi.org/10.1017/CBO9780511813085

Chen, X., Shin, G.-H., & Lee, J. (2025). Exploring metapragmatics of politeness lexemes using a computational approach. *Journal of Politeness Research*, *21*(1), 167–192. https://doi.org/10.1515/pr-2023-0021

Danescu-Niculescu-Mizil, C., Sudhof, M., Jurafsky, D., Leskovec, J., & Potts, C. (2013). A computational approach to politeness with application to social factors. In H. Schuetze, P. Fung, & M. Poesio (Eds.), *Proceedings of the 51st Annual Meeting of the Association for Computational Linguistics (Volume 1: Long Papers)* (pp. 250-259). Association for Computational Linguistics. https://aclanthology.org/P13-1025/

Darwish, K., Habash, N., Abbas, M., Al-Khalifa, H., Al-Natsheh, H. T., Bouamor, H., Bouzoubaa, K., Cavalli-Sforza, V., El-Beltagy, S. R., El-Hajj, W., Jarrar, M., & Mubarak, H. (2021). A panoramic survey of natural language processing in the Arab world. *Commun. ACM*, *64*(4), 72–81. https://doi.org/10.1145/3447735

Lakoff, R. (1973). Language and woman's place. *Language in Society*, *2*(1), 45–79. https://doi.org/10.1017/S0047404500000051

Landis, J. R., & Koch, G. G. (1977). The measurement of observer agreement for categorical data. *Biometrics*, *33*(1), 159–174.

Leech, G. N. (2016). *Principles of Pragmatics*. Routledge. https://doi.org/10.4324/9781315835976

Priya, P., Firdaus, M., & Ekbal, A. (2024). Computational Politeness in Natural Language Processing: A Survey. *ACM Comput. Surv.*, *56*(9), 241:1-241:42. https://doi.org/10.1145/3654660

Srinivasan, A., & Choi, E. (2022). TyDiP: A Dataset for Politeness Classification in Nine Typologically Diverse Languages. In Y. Goldberg, Z. Kozareva, & Y. Zhang (Eds.), *Findings of the Association for Computational Linguistics: EMNLP 2022* (pp. 5723–5738). Association for Computational Linguistics. https://doi.org/10.18653/v1/2022.findings-emnlp.420

Yin, Z., Wang, H., Horio, K., Kawahara, D., & Sekine, S. (2024). Should We Respect LLMs? A Cross-Lingual Study on the Influence of Prompt Politeness on LLM Performance. In J. Hale, K. Chawla, & M. Garg (Eds.), *Proceedings of the Second Workshop on Social Influence in Conversations (SICon 2024)* (pp. 9–35). Association for Computational Linguistics. https://doi.org/10.18653/v1/2024.sicon-1.2